\documentclass[UTF8]{article}
\usepackage[colorlinks,
            linkcolor=blue,
            anchorcolor=blue,
            citecolor=green]{hyperref}
\usepackage{spconf,amsmath,graphicx}
\usepackage{amsfonts}
\usepackage{color}
\usepackage{multirow}
\usepackage{pifont}
\usepackage{makecell}
\usepackage{float}
\usepackage[marginal]{footmisc}
\usepackage[numbers,sort&compress]{natbib}
\begin{document}
%\ninept

\title{focus on semantic consistency for cross-domain crowd understanding}
\name{Tao Han, Junyu Gao, Yuan Yuan, Qi Wang* \footnotemark[1]}

\address{School of Computer Scinence and Center for OPTical IMagery Analysis and Learning(OPTIMAL),\\ Northwestern Polytechnical University, Xi'an 710072, Shaanxi, P.R.China}

\maketitle
\renewcommand{\thefootnote}{\fnsymbol{footnote}}
\footnotetext[1]{Qi Wang is the corresponding author. \quad This work was supported by the National Key R\&D Program of China under Grant 2017YFB1002202, National Natural Science Foundation of China under Grant U1864204, 61773316, 61632018 and 61825603.}

\begin{abstract}
For pixel-level crowd understanding, it is time-consuming and laborious in data collection and annotation. Some domain adaptation algorithms try to liberate it by training models with synthetic data, and the results in some recent works have proved the feasibility. However, we found that a mass of estimation errors in the background areas impede the performance of the existing methods. In this paper, we propose a domain adaptation method to eliminate it. According to the semantic consistency, a similar distribution in deep layer's features of the synthetic and real-world crowd area, we first introduce a semantic extractor to effectively distinguish crowd and background in high-level semantic information. Besides, to further enhance the adapted model, we adopt adversarial learning to align features in the semantic space. Experiments on three representative real datasets show that the proposed domain adaptation scheme achieves the state-of-the-art for cross-domain counting problems.%\footnotemark[2]
\end{abstract}
%\footnotetext[2]{Code: \url{https://github.com/taohan10200/cross-domain-CC}}
\begin{keywords}
Crowd counting, domain adaptation, semantic consistency, adversarial learning
\end{keywords}
\section{Introduction}
\label{sec:intro}
Over the past few years, crowd understanding has become increasingly influential in the field of computer vision. Because it plays an important role in social management, such as video surveillance, public area planning, crowd congestion warning, traffic flow monitoring and so on \cite{C1, C5, C30}. Crowd counting is a foundation of crowd understanding, this task strives to understand people from images, predict density maps and estimate the number of pedestrians for crowd scenes. At present, many CNN-based approaches \cite{C2, C3, C27, C18,gao2019pcc} have achieved phenomenal performance, and their success is driven by the availability of public crowd datasets. Unfortunately, the existing datasets (such as Shanghai Tech A/B \cite{C7}, and UCF-QNRF \cite{C8}, etc.) are so small-scale that makes it difficult for trained models to perform well in other scenarios. The high dependency on annotated data makes it difficult to deploy to social management.

 To address the problem of data scarcity in crowd counting, some works \cite{C19, C20, C21} explore unsupervised or weakly supervised crowd counting, but these methods did not completely get rid of the dependence on manually annotated data. Inspired by the application of synthetic data \cite{C22, C23} in other visual fields, a large-scale crowd counting dataset named GCC was established by Wang \emph{et al.} \cite{C9}, where the data is generated and annotated automatically by a computer game mod. Although this novel data generation method solves the challenge of manually labeling data, one problem that comes with it is that the gap between the synthetic scenes and the real-world scenes is too large. Therefore, a well-trained model on the GCC dataset doesn't work well in the real world. Some of the recent work \cite{C13, C14, C15,Lee_2019_CVPR,Pan_2019_CVPR} provided us with a domain adaptation strategy, using image style transfer networks to narrow the domain gap. For the cross-domain crowd counting problem, \cite{C9} also proposed a domain adaptation method to make synthetic data closer to real data in visual perception via the SE Cycle GAN \cite{C9} network. This method of using image style transfer has achieved better results than the CNN models without domain adaptation, but the regression error in the background area reduces counting performance.
\begin{figure*}[htbp]
  \centering
  \centerline{\includegraphics[width=17cm]{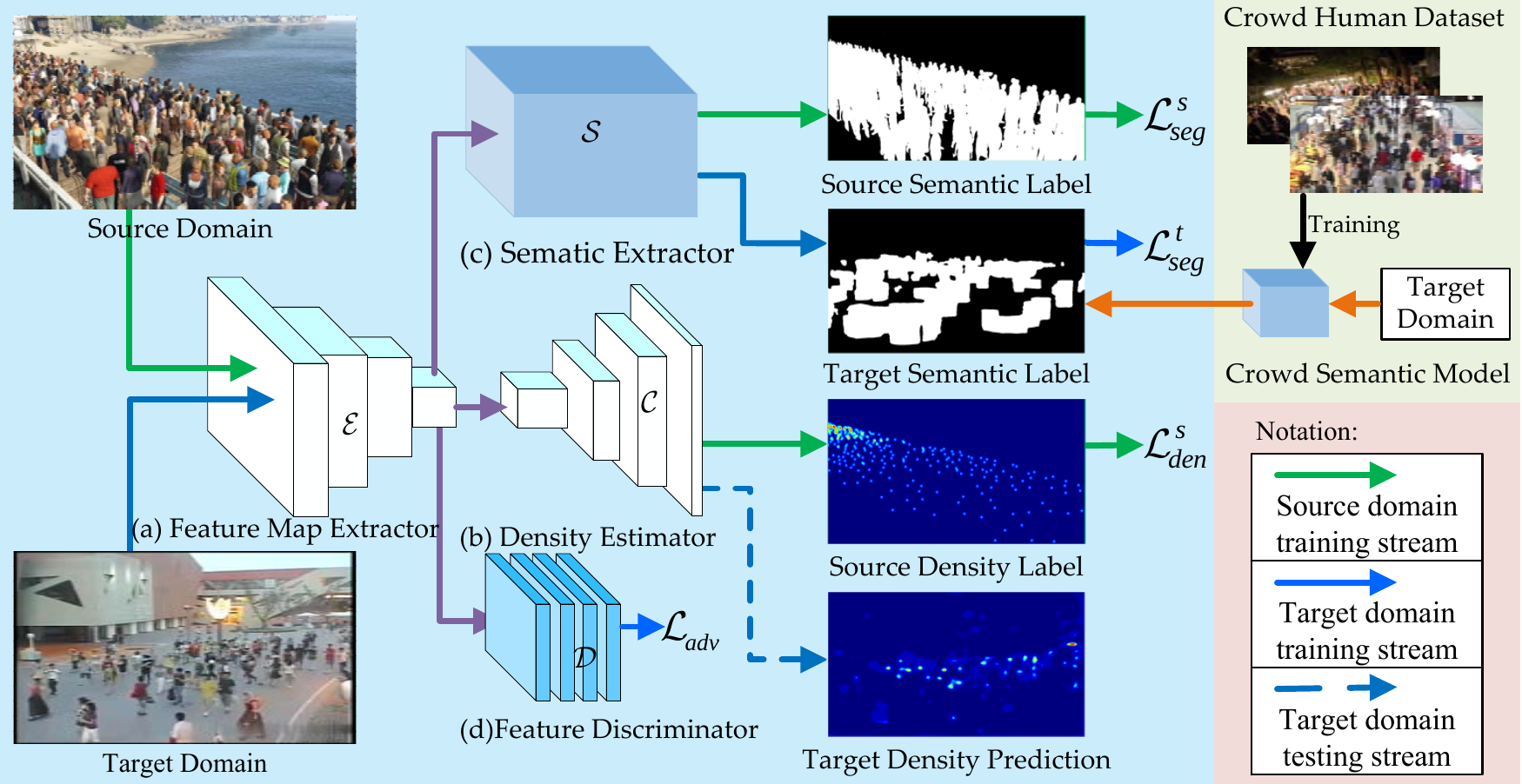}}

 %\centerline{(a) Result 1}\medskip

\caption{The architecture of the proposed domain adaptation network for across-domain crowd understanding.}
\label{fig:framework}
 \vspace{-0.5cm}
 %\hfill
\end{figure*}

In this paper, the central issue is \emph{how to design a better domain adaptation scheme for reducing the counting noise in the background area.} After observing a large number of images in the GCC \cite{C9} dataset, we find that the characters in the synthetic scene and the real scene have a high degree of visual similarity, while the background has a large gap. This similarity is more obvious in the high-level semantic representation. We assume that if the network can pay more attention to the semantic consistency of the crowd, it will help to narrow the domain gap. To make our adapted model can extract the semantic consistency feature for synthetic and real-world data, we first introduce a semantic extractor by further exploiting the semantic label. Considering that GCC \cite{C9} dataset provides the mask for crowd area, correspondingly, we train a segmentation model and get the semantic label for real datasets in free. Furthermore, we adopt adversarial learning to align the semantic space. Based on the above two methods, our domain adaption framework is formed. A detailed description of the network will be in the second \ref{sec:framework}.

In summary, our main contributions of this paper are as follows:
\begin{itemize}
\setlength{\itemsep}{0pt}
\setlength{\parsep}{0pt}
\setlength{\parskip}{0pt}
  \item  We exploit a large-scale human detection dataset to train a crowd semantic model, which can generate crowd semantic labels for all real crowd datasets.
  \item  We propose a domain adaptive network based on semantic consistency, which strives to focus on the consistent feature of the cross-domain crowd.
  \item  We apply our framework to three real datasets, and it yields a new record on the across-domain crowd counting problem.
\end{itemize}

\section{Proposed Framework}
\label{sec:framework}
The overall architecture of our proposed domain adaptation framework is illustrated in Fig.\ref{fig:framework}. In this section, we first describe the details of the architecture, then introduce the various loss functions, and finally, we give the training details of the framework.
\vspace{-10pt}
\subsection{Framework Details}
\label{subsec:framework_dtail}
For the sake of understanding, some symbolic definitions are given here. In this paper, the number of available annotation data is $N_{s}$ source domain images and $N_{t}$ target domain images, denoted as $X^{s}\!=\!\left\{\left(x_{i}^{s}, y_{i}^{s},z_{i}^{s}\right)\right\}_{i=1}^{N_{s}}$ and $X^{t}\!=\!\left\{\left(x_{i}^{t}, z_{i}^{t}\right)\right\}_{i=1}^{N_{t}}$ respectively. Each image $x_{i}^{s},x_{i}^{t} \in \mathbb{R}^{H \times W \times 3}$ has RGB three-channel pixels with height $H$, width $W$. The source domain image $x_{i}^{s}$ has one-channel per-pixel label of head position $y_{i}^{s} \in \{0,1\}^{H \times W }$ and crowd mask $z_{i}^{s}\in\{0,1\}^{H \times W}$, while the target image $x_{i}^{t}$ only has a roughly crowd mask $z_{i}^{t}\in\{0,1\}^{H \times W}$. As shown in Fig. \ref{fig:framework} (a), (b), (c) and (d), there are four sub-networks, namely, the common feature map extractor $\mathcal{E}$, the crowd density estimator $\mathcal{C}$, the crowd semantic extractor $\mathcal{S}$, and the feature discriminator $\mathcal{D}$. They are parameterized by $\theta_{e}$, $\theta_{c}$, $\theta_{s}$ and $\theta_{d}$, respectively.

\textbf{Feature Map Extractor}\quad Considering that the VGG network has less computation and strong feature representation ability, we choose a pre-trained VGG16 \cite{C16} with batch normalization as the frond-end feature map extractor. It is also fair to compare with the previous work. Given a pair of images $x^{s}$ and $x^{t}$ as input, the output produced by the feature extractor $\mathcal{E}$ can be represented by the following mapping:
\begin{equation}
\setlength{\abovedisplayskip}{3pt}
\setlength{\belowdisplayskip}{3pt}
%\left\{
%\begin{aligned}
f^{s}=\mathcal{E}\left(x^{s} ; \theta_{e}\right),
f^{t}=\mathcal{E}\left(x^{t} ; \theta_{e}\right).
%\end{aligned}
%\right\}
\end{equation}

\textbf{Semantic Extractor(SE)}\quad The semantic extractor is designed to predict the crowd mask. We propose a crowd semantic extractor by slightly modifying the pyramid module of PSPNet proposed by Zhao \emph{et al.} \cite{C12}, which fuses features under four different pyramid scales to aggregate context information for different regions. The fusion feature outputs the final semantic prediction $\hat{z}$  through the module, which is defined as follows:
\begin{equation}
\setlength{\abovedisplayskip}{3pt}
\setlength{\belowdisplayskip}{3pt}
%\left\{
%\begin{aligned}
\hat{z}^{s}=\mathcal{S}\left(f^{s} ; \theta_{s}\right),
\hat{z}^{t}=\mathcal{S}\left(f^{t} ; \theta_{s}\right).
%\end{aligned}
%\right\}
\end{equation}

\textbf{Feature Discriminator(FD)}\quad The feature discriminator achieves domain adaptation by identifying whether the input features $f$ come from the source domain or the target domain. We use an architecture similar to \cite{C29}. The network consists of $4$ convolution layers with kernel $4\times4$ and stride of $2$. Given the features $f^{s}$ and $f^{t}$, we forward $f^{s}$ and $f^{t}$ to a feature discriminator $\mathcal{D}$. The discriminant result $p^{s}$, $p^{t}$ can be described by the following mapping:
\begin{equation}
\setlength{\abovedisplayskip}{3pt}
\setlength{\belowdisplayskip}{3pt}
p^{s}=\mathcal{D}\left(f^{s} ; \theta_{d}\right),
p^{t}=\mathcal{D}\left(f^{t} ; \theta_{d}\right).
\end{equation}

\textbf{Density Estimator}\quad Since the core of this paper is not the design of crowd density predictor, we only adopt a series of simple convolution and up-sample layers to build our crowd density predictor. As shown in Fig. \ref{fig:framework} (b), each convolution $3\times3$ layer is followed by a up-sample layer. Finally, the source domain density map participating in the training is defined as:
\begin{equation}\label{equ:3}
\setlength{\abovedisplayskip}{3pt}
\setlength{\belowdisplayskip}{3pt}
\begin{aligned}
\hat{y}^{s}=\mathcal{C}\left(f^{s} ; \theta_{c}\right).
\end{aligned}
\end{equation}
\subsection{Loss Functions}
In section \ref{subsec:framework_dtail}, we define the outputs $\left\{\hat{y}^{s} ,\hat{y}^{t}\right\}$, $\left\{p^{t}, p^{t}\right\}$ and $\hat{z}^{s}$. Correspondingly, we define some loss functions to train our model. The training of the proposed framework is to minimize a weighted combination loss function with respect to the parameters $\left\{\theta_{e},\theta_{s},\theta_{c},\theta_{d}\right\}$ of the sub-networks. The final objective function is summed as:
\begin{equation}
\small
\setlength{\abovedisplayskip}{3pt}
\setlength{\belowdisplayskip}{3pt}
\begin{array}{r}{\mathcal{L}\left(X^{s},X^{t} ; \theta_{e}, \theta_{s}, \theta_{c}\right)\!=\! \mathcal{L}_{den}^{s}\!+\!\lambda^{s}\!\mathcal{L}_{seg}^{s}\!+\!\lambda^{t}\! \mathcal{L}_{seg}^{t}\!+\!\lambda^{d}\!\mathcal{L}_{adv}}\end{array}\!,
\end{equation}
where $\lambda^{s}$, $\lambda^{t}$ denote the weighting parameters of the
different mask segmentation loss functions. Since the loss of mask segmentation is designed to assist the network to focus on semantic consistency of the crowd area, its weight should be set carefully. After the experimental testing, we found that it better to set them all at 0.01. $\lambda^{d}$ is chosen empirically to strike a balance among the model capacity,In the following, we elaborate on each of these loss functions.

\textbf{Crowd Density Estimation Loss}\quad To predict density maps $\hat{y}^{s}$ for source-domain images $x^{s}$, the density estimation loss $\mathcal{L}_{den}^{s}$ given by the typical Euclidean distance based on the source domain ground truths $y^{s}$ is to supervised train the feature extractor $\mathcal{E}$ and the crowd density predictor $\mathcal{C}$.In symbols, it is defined as follows:
\begin{equation}
\setlength{\abovedisplayskip}{3pt}
\setlength{\belowdisplayskip}{3pt}
\mathcal{L}_{d e n}^{s}\left(\theta_{e},\theta_{c}\right)=\frac{1}{2N} \sum_{i=1}^{N}\left\|\hat{y}_{i}^{s}-F\left(y_{i}^{s}\right)\right\|^{2},
\end{equation}
where $F\left(y_{i}^{s}\right)$ is the gaussian density map of ${y}_{i}^{s}$, which is generated following a  gaussian kernel function of \cite{C2}. $\hat{y}_{i}^{s}$ is the estimated density map of $x_{i}^{s}$, which comes from the mapping defined in Equ. \ref{equ:3}.

\textbf{Crowd Semantic Segmentation Loss}\quad The $\mathcal{L}_{seg}$ is designed as a auxiliary loss to extractor the semantic consistency feature for different domain's crowd. The goal of the semantic extractor is to learn correspondingly from the input characteristics $f^{s},f^{t}$ to predict the crowd mask $z^{s},z^{t}$. To train the network, we first introduce the source domain crowd segmentation loss functions $\mathcal{L}_{seg}^{s}$, which is a binary class entropy, defined as:
\begin{equation}
\setlength{\abovedisplayskip}{0pt}
\setlength{\belowdisplayskip}{0pt}
\small
\mathcal{L}_{seg}^{s}\!\left(\theta_{e},\theta_{s}\right)\!=\!-\frac{1}{N}\sum_{i=1}^{N}\!\left(z_{i}^{s} \log \hat{z}_{i}^{s}+\left(1-z_{i}^{s}\right) \log \left(1-\hat{z}_{i}^{s}\right)\right),
\end{equation}
where $z_{i}^{s}$ is the source domain crowd semantic label. $\hat{z}_{i}^{s}$ is the probability of each pixel in the semantic prediction map activated by a sigmoid function.

 As mentioned in section \ref{sec:intro}, we train a model with the CrowdHuman \cite{C10} dataset to generate the semantic label $z^{t}$ for free. Since the center of this paper is not on how to acquire the semantic label, it would not be elaborated here due to the limited space. The visualization of $z^{t}$ can be seen in figure \ref{fig:framework}. It can be observed that the target domain semantic label $z^{t}$ is not as reliable as the source domain $z^{s}$ because the pedestrian label is rectangular in object detection. To eliminate the negative effects of such inaccurate segmentation, we only use background labels to promote training, that is, we ignore the white region of the semantic lable $z^{t}$. Mathematically, we designed Equ. \ref{equ:8} to filter the prediction mask $\hat{z}^{t}$.
 \vspace{-0.3cm}
\begin{equation}\label{equ:8}
\setlength{\abovedisplayskip}{3pt}
\setlength{\belowdisplayskip}{3pt}
H(\hat{z}^{t},z^{t})=\hat{z}^{t}\left(1-z^{t}\right)+z^{t},
\end{equation}
 \vspace{-0.2cm}
let $\overline{z}^{t}=H(\hat{z}^{t},z^{t})$, we define loss function $\mathcal{L}_{seg}^{t}$ as follow:
\begin{equation}
\small
\setlength{\abovedisplayskip}{3pt}
\setlength{\belowdisplayskip}{3pt}
\begin{aligned}
\mathcal{L}_{seg}^{t}\!\left(\theta_{e},\theta_{s}\right)\!=\!-\frac{1}{N}\!\sum_{i=1}^{N}\left(z_{i}^{t} \log \overline{z}_{i}^{t} + \left(1\!-\!z_{i}^{t}\right) \log\left(1\!-\!\overline{z}_{i}^{t}\right)\right).
\end{aligned}
\end{equation}

\renewcommand{\multirowsetup}{\centering}
\begin{table*}[ht]
\setlength{\belowdisplayskip}{1pt}
\centering
    \caption{The performance of No Adaptation, Cycle GAN, SE Cycle GAN
     and our approaches on the three real-world datasets.}\label{Tab:metric}
   % \scriptsize
    %\resizebox{\textwidth}{12mm}{
    \setlength{\tabcolsep}{1.2mm}{
    \begin{tabular}{c|c|c|c|c|c|c|c|c|c|c|c|c|c}
    \Xhline{1.2pt}
    \multirow{2}{*}{Method} &\multirow{2}{*}{DA}&\multicolumn{4}{c|}{ShanghaiTech Part A}& \multicolumn{4}{c|}{ShanghaiTech Part B} &\multicolumn{4}{c}{UCF-QNRF}\\
    \cline{3-14}
    && MAE & MSE & PSNR & SSIM & MAE & MSE & PSNR & SSIM & MAE & MSE & PSNR & SSIM\\
    \hline
    NoAdpt \cite{C9}      &\ding{55} &160.0&216.5&19.01 &0.359 & 22.8 & 30.6 &24.66 &0.715 & 275.5 &458.5 & 20.12 & 0.554 \\
    \hline
    Cycle GAN \cite{C17}   &\ding{52} &143.3&204.3&19.27&0.379& 25.4 & 39.7 &24.60 &0.763 & 257.3 &400.6 & 20.80 & 0.480 \\
    \hline
    SE Cycle GAN \cite{C9}&\ding{52}&\bfseries123.4&193.4&18.61&0.407 & 19.9 & 28.3 &24.78 &0.765 & 230.4 &\bfseries384.5 & 21.03 & 0.660 \\
    \Xhline{1.2pt}
    NoAdpt(ours)&\ding{55} & 190.8 & 298.1 &20.57& 0.457 &24.6& 33.7 &24.14 &0.687 & 296.1 &467.9 & 20.46 & 0.513 \\
    \hline
    SE+FD     &\ding{52} &129.3& \bfseries187.6 &\bfseries21.58&\bfseries0.513&\bfseries 16.9 & \bfseries24.7 &\bfseries26.20 &\bfseries0.818 & \bfseries221.2 &390.2 & \bfseries23.10 & \bfseries0.708 \\
    \Xhline{1.2pt}
    \end{tabular}}
\end{table*}
\textbf{Semantic Space Adversarial Loss}\quad In the hopes of making the $\mathcal{E}$ extract consistent features for source domain and target domain, we introduce an adversarial loss $\mathcal{L}_{adv}$ following \cite{C29}. Specifically, we first train a discriminator $\mathcal{D}$ to distinguish between the source fearure $f^{s}$ and target feature $f^{t}$ by minimizing a supervised domain loss. (i.e. $\mathcal{D}$ should ideally output $P^{s}$ to 1 in the source feature $f^{s}$ and  $p^{t}$ to 0 for that in the target feature $f^{t}$). We then update the $\mathcal{E}$ to fool the discriminator $\mathcal{D}$ by inverting its output $p^{t}$ from 0 to 1, that is, by minimizing
 \vspace{-0.2cm}
\begin{equation}
\setlength{\abovedisplayskip}{3pt}
\setlength{\belowdisplayskip}{3pt}
\mathcal{L}_{a d v}\left(\theta_{e}\right)=-\sum_{h, w} \log\left(p^{t}\right),
\end{equation}
where $h$ and $w$ are the coordinate dimensions of the output $p^{t}$.
\vspace{-0.25cm}
\subsection{Implementation}
\vspace{-0.05cm}
\textbf{Scene Regularization}\quad GCC \cite{C9} is a large-scale synthetic dataset that includes a variety of scenarios, weather, and time periods.  Therefore, adding the entire dataset to the training will bring negative effects. To eliminate this adverse effect and facilitate fair comparison, we adopted the Scene Regularization \cite{C9} to select the images.

\noindent
\textbf{Training Details}\quad During the training phase, the goal is to optimize $\mathcal{E}$, $\mathcal{S}$, and $\mathcal{C}$. Due to limited memory, we set the batch size to $8$ and perform randomly cropping the images with a size of $480\times640$. and we adopt the Adam algorithm to optimize the network, the initialization learning rate is  set to $10^{-5}$. The training and evaluation are performed on NVIDIA GTX $1080$Ti GPU using the $C^{3}$ framework \cite{C28}.

\section{Experimental Results}
\subsection{ Evaluation Metrics}
\vspace{-0.1cm}
\textbf{Count Error}\quad Following the convention of existing works \cite{C25,C26}, we adopt Mean Absolute Error (MAE) and Mean Squared Error (MSE)  as count error evaluation metrics, which are formulated as below:
\begin{equation}
\setlength{\abovedisplayskip}{3pt}
\setlength{\belowdisplayskip}{3pt}
\small
%\left\{
MAE=\frac{1}{N} \sum_{i=1}^{N}\left|y_{i}-\hat{y}_{i}\right|,
MSE=\sqrt{\frac{1}{N} \sum_{i=1}^{N}\left|y_{i}-\hat{y}_{i}\right|^{2}},
%\right\}
\end{equation}
where $N$ is the number of testing images, $y_{i}$ is the ground truth counting value and $\hat{y}_{i}$ is the estimated counting value for the $i$th test image.

\noindent
\textbf{Density Map Quality}\quad Besides, to evaluate the quality of the predicted density maps, we also introduce  PSNR (Peak Signal-to-Noise Ratio) and SSIM (Structural Similarity in Image)\cite{C11}.
\vspace{-1pt}
\subsection{Datasets}
We conduct the experiments on the ShanghaiTech PartA/B dataset \cite{C7} and UCF-QNRF dataset \cite{C8} . The ShanghaiTech Part A contains 482 crowd images (300 training and 182 testing images), and the average number of the pedestrian is 501. The ShanghaiTech Part B is with $716$ images ($400$ training and $316$ testing images),and the average number of people per image is about $123$. The UCF-QNRF is a congested crowd dataset, which consists of $1,535$ images($1201$ training and $334$ testing images), with the count ranging from $49$ to $12,865$, and the average number of the pedestrian is 815 per image.
\begin{figure}[ht]
\setlength{\abovedisplayskip}{3pt}
\setlength{\belowdisplayskip}{3pt}
\begin{minipage}[b]{1.0\linewidth}
  \centering
  \centerline{\includegraphics[width=8.5cm]{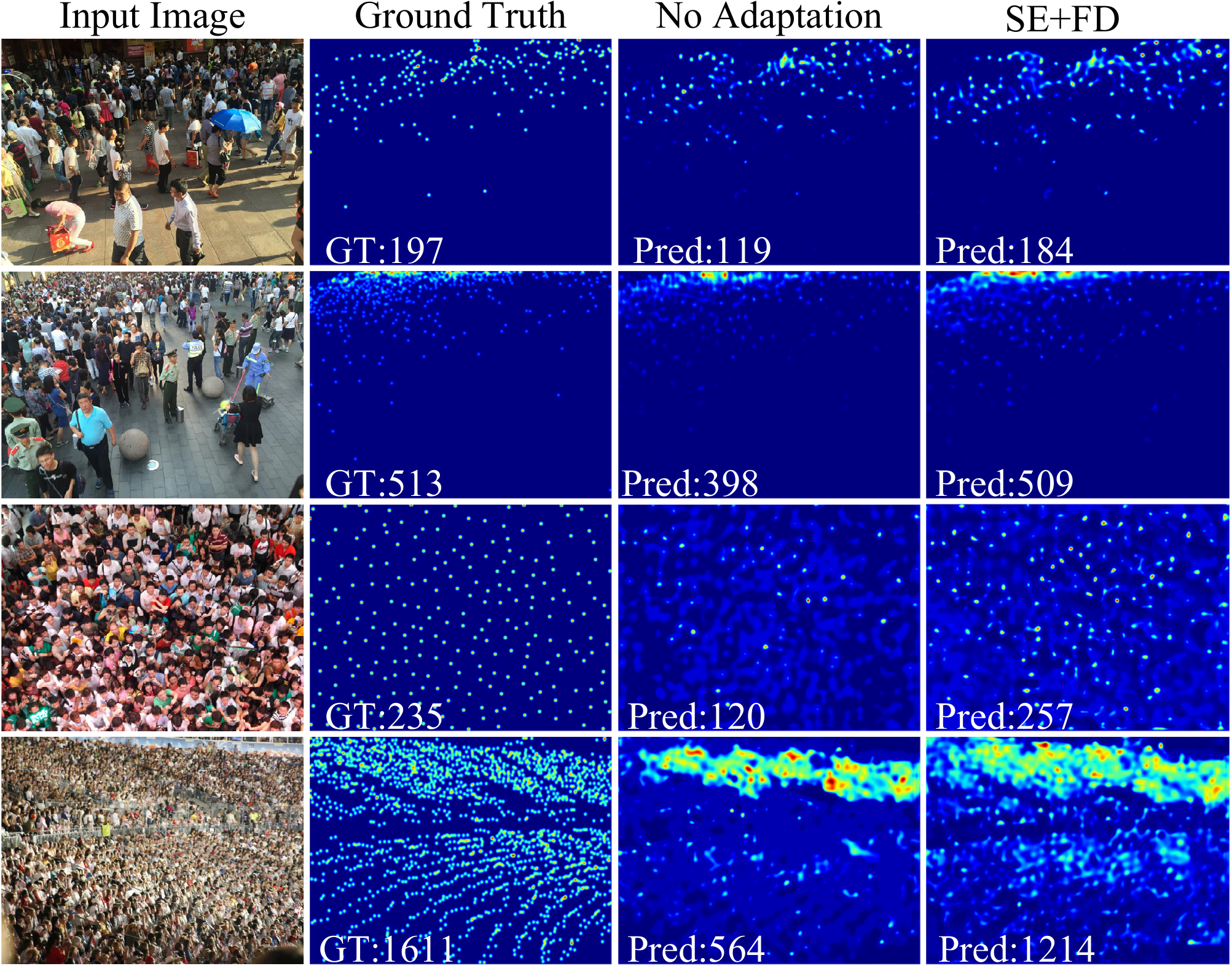}}
%  \vspace{2.0cm}
  %\centerline{(a) Result 1}\medskip
\end{minipage}
%\hfill

\caption{Exemplar results of adaptation from GCC to Shanghai Tech Part B and UCF-QNRF datasets.}
\label{fig:result}
\end{figure}

\subsection{Performance Comparison}
In this section, we perform experiments on three typical datasets for across domain crowd counting and then compare the performance of our proposed method with the state-of-the-art SE Cycle GAN \cite{C9}.
The test results are shown in Table \ref{Tab:metric}. Compared with Cycle GAN \cite{C17} and SE Cycle GAN \cite{C9}, which adopt image style transfer, our approach is more practical and yields better results. For Shanghai B, with no domain adaptation, the results of our baseline model are not as good as the SFCN \cite{C9}, because we adopt a simpler crowd counter. However, After the domain adaptation, our model achieves comprehensive transcendence in all metrics. Our proposed method reduces the MAE to $16.9$ and the MSE to $24.7$, which dropped by \textcolor{red}{$31.3\%$} and \textcolor{red}{$26.7\%$} compared with the baseline model. In terms of image quality, we also obtain better SSIM and PSNR. For the Shanghai B and UCS-QNRF datasets, both are the congested datasets, our proposed method also performs a better result than the state-of-the-art SE Cycle GAN\cite{C9}.

Fig.\ref{fig:result} shows the visualization results on the real datasets. It can be observed visually that the Column 4 with domain adaptation is closer to the ground truth than Column 3 without domain adaptation in terms of image quality and crowd counting. This improvement is because our method focuses on the consistency of crowd features, thus reducing the estimation error of background.
\vspace{-1pt}
\section{Conclusion}
\vspace{-6pt}
In this paper, we aim to count people for real scenarios using synthetic datasets. For the problem of background estimation error in the existing methods, we propose an effective domain adaptation framework, which emphasizes the network to concentrate on the crowd-area semantic consistency of the source domain and target domain by using two adaptation strategies. Experiments on high-density and low-density datasets show that our proposed method achieves state-of-the-art performance. In future work, we will further utilize high-level semantic information and domain transfer to achieve higher precision crowd counting.

\vfill\pagebreak

\bibliographystyle{IEEEbib}
\bibliography{strings,refs}

\end{document}